\documentclass[10pt,twocolumn,letterpaper]{article}

\usepackage[pagenumbers]{cvpr} 
\usepackage{graphicx}
\usepackage{amsmath}
\usepackage{amssymb}
\usepackage{diagbox}
\usepackage{ctable}
\usepackage{booktabs}
\usepackage{mathrsfs}
\usepackage{changepage}
\usepackage{multirow}
\usepackage{adjustbox}
\usepackage{makecell}
\usepackage{esvect}
\usepackage{bbding}
\usepackage{pifont}
\usepackage{wasysym}
\usepackage{amssymb}
\usepackage{color, colortbl}
\definecolor{gray}{gray}{0.85}

\definecolor{cvprblue}{rgb}{0.21,0.49,0.74}
\usepackage[pagebackref,breaklinks,colorlinks,citecolor=cvprblue]{hyperref}

\def\paperID{4} 
\def\confName{CVPR Workshop}
\def\confYear{2024}

\title{MMCBE: Multi-modality Dataset for Crop Biomass Prediction and Beyond}

\author{
Xuesong Li$^{1}$, 
Zeeshan Hayder$^{1}$, 
Ali Zia$^{1}$, 
Connor Cassidy$^{1}$, 
Shiming Liu$^{1}$, 
Warwick Stiller$^{1}$, \\
Eric Stone$^{2}$, 
Warren Conaty$^{1}$, 
Lars Petersson$^{1}$, 
Vivien Rolland$^{1}$
 \\
$^{1}$CSIRO, $^{2}$Austraila National University, Australia \\ 
{\tt\small xuesong.li@csiro.au} \\ 
}

\begin{document}
\maketitle


\begin{abstract}

Crop biomass, a critical indicator of plant growth, health, and productivity, is invaluable for crop breeding programs and agronomic research. However, the accurate and scalable quantification of crop biomass remains inaccessible due to limitations in existing measurement methods. One of the obstacles impeding the advancement of current crop biomass prediction methodologies is the scarcity of publicly available datasets. Addressing this gap, we introduce a new dataset in this domain, i.e. Multi-modality dataset for crop biomass estimation (MMCBE). Comprising 216 sets of multi-view drone images, coupled with LiDAR point clouds, and hand-labelled ground truth, MMCBE represents the first multi-modality one in the field. This dataset aims to establish benchmark methods for crop biomass quantification and foster the development of vision-based approaches. We have rigorously evaluated state-of-the-art crop biomass estimation methods using MMCBE and ventured into additional potential applications, such as 3D crop reconstruction from drone imagery and novel-view rendering. With this publication, we are making our comprehensive dataset available to the broader community.

\end{abstract}    
\section{Introduction}
\label{sec:intro}

The development of advanced agricultural technologies is required to address the challenges of food security, which are intensified by a rapidly growing global population. Among the various traits monitored in sustainable agricultural systems, biomass (defined as the amount of organic matter produced by plants) is pivotal, as monitoring biomass aids in assessing the success of plant establishment and informs critical decisions regarding replanting or the application of treatments such as chemical sprays.  The estimation and mapping of biomass on a large scale, including up to global maps, have been extensively explored within the remote sensing community~\cite{DUNCANSON2022112845, bullock2023estimating}. These large-scale biomass estimates, primarily focused on vegetation biomass including forests, are typically conducted at resolutions ranging from 30 meters to 1 kilometre. However, this paper narrows its focus to the high-resolution estimation of crop biomass, and the term `biomass' specifically refers to above-ground biomass (i.e. excluding roots) for agricultural crops, which is crucial for the precision required in contemporary agricultural monitoring systems. Accurate and automated estimation of crop biomass is a complex task that necessitates the integration of computer vision, robotics, and machine learning technologies. Despite the critical importance of this parameter, accurate and scalable measurement of crop biomass remains a challenge due to the limitations of existing methodologies~\cite{pan2022biomass, loudermilk2009ground}.

Traditional methods for quantifying crop biomass typically involve the harvesting and measuring of biomass within a small, designated plot area (e.g., a square meter or a linear meter). These values are then extrapolated to estimate the biomass for an entire plot or field. Such approaches are inherently destructive and labour-intensive, limiting their scalability and eliminating the generation of temporal data on biomass changes over time. In contrast, advancements in computer vision technology present a non-destructive alternative, enabling the estimation of biomass across different time points~\cite{walter2019estimating, jimenez2018high, liu2018robust, caballer2021prediction, sun2018field, polo2009tractor, li2020real, ten2019biomass, pan2022biomass}. Specifically, Light Detection and Ranging (LiDAR) technology has been employed to capture critical plant metrics, such as height and point density, which serve as proxies for biomass estimation. While these proxy-based methods offer a degree of generalizability, their reliance on simplistic 3D feature representations and the challenge of canopy self-occlusion have historically limited their accuracy. To address these shortcomings, deep learning techniques have been integrated into the analysis of point clouds, significantly enhancing the precision of biomass estimations by overcoming the issue of self-occlusion and extracting more robust 3D features~\cite{pan2022biomass, ma2023automated}. This indicates that an increasing number of deep-learning-based approaches will might be developed for crop biomass prediction tasks.


The significance of comprehensive datasets, such as ImageNet~\cite{deng2009imagenet} and CoCo~\cite{lin2014microsoft}), to the advancement of convolutional neural networks (CNN) in image recognition tasks has been well-documented~\cite{ren2015faster, li2019detection, li2023efficient, resnet_simple}. An important driving force behind the development of deep learning methodologies is the availability of publicly accessible datasets, which not only offer expansive training data but also establish transparent benchmarks for evaluating performance. To further enhance deep-learning-based approaches for crop biomass prediction, it is critical to increase the availability of public-access datasets within the agricultural technology communities. However, despite the urgent need for such datasets, there is a notable scarcity in the field. To the best of our knowledge, there is only one publicly available crop dataset dedicated to LiDAR-based biomass estimation ~\cite{pan2022biomass}. To address this gap, we are contributing a new dataset for crop biomass prediction tasks, i.e. MMCBE, and it includes both point cloud and multi-view images with manually measured biomass ground truth, representing the first multi-modality dataset in this area. Our point cloud dataset can expand the training dataset for biomass prediction approaches and evaluate current methods. The multi-view image dataset, the first in this domain, aims to inspire further vision-based biomass prediction methods. These methods are particularly promising due to the widespread availability and affordability of RGB cameras and have the potential for large-scale application. Our main contributions in this paper are listed as follows:

\begin{itemize}
  \item We publish the first multi-modality dataset (i.e. MMCBE) which includes a set of 216 multi-view drone image sets with LiDAR point cloud and hand-labelled ground truth. The dataset will be released here, \url{https://github.com/Benzlxs/MMCBE}
  \item We provide a benchmark for quantifying crop biomass estimation approaches.
  \item We explored additional potential tasks, such as 3D crop reconstruction from multi-view images and novel-view synthesis.
\end{itemize}

\noindent In the rest of the paper, section~\ref{sec:related_work} will present current main biomass prediction methods and other relevant datasets in this area, followed by the introduction about our multi-modality dataset and data processing pipeline in section~\ref{sec:dataset}. We will evaluate state-of-the-art methods and explore other computer vision tasks on our dataset in section~\ref{sec:experiments}, and then finish the paper with section~\ref{sec:conclusion}.

\section{Related work}
\label{sec:related_work}
\subsection{Biomass estimation}
Crop biomass estimation approaches are primarily categorized into traditional and deep-learning-based methods. Traditional approaches refer to simple models with fewer learning parameters, such as linear regression and random forest. In contrast, deep-learning-based methods involve complex neural networks with millions of parameters, such as Transformer and CNN framework~\cite{vaswani2017attention, resnet_simple, li2019three}.

\subsubsection{Traditional biomass estimation}
The manual biomass measurement method involves cutting above-ground plants of a small sampled area, drying them in an oven, and weighing the dry materials. This conventional method is labour-intensive, destructive, and prone to approximation error~\cite{Zhao2021,CATCHPOLE1992}. Due to this, there has been a growing interest in developing automated approaches for biomass estimation.

The adoption of canopy height or density as a surrogate for biomass estimation has gained considerable traction in recent literature~\cite{ loudermilk2009ground, Gebbers2011, tilly2014multitemporal, shu2023using}. Notably, Saeys et al.~\cite{saeys2009estimation} employed statistical models to estimate crop density through two LiDAR scanning frequencies. Tilly et al.~\cite{ tilly2014multitemporal} augmented their platform with a field spectrometer, leveraging 3D and spectral data fusion to derive bivariate biomass regression models~\cite{ tilly2015fusion}. Li et al.~\cite{li2015airborne} employed airborne LiDAR alongside Pearson's correlation analysis and structural equation modeling) to gauge plant height and leaf area index. Shu et al.~\cite{shu2023using} used high-resolution unmanned aerial vehicle (UAV) cameras to measure the canopy height and predicted the biomass with a linear regression model. While several studies have explored alternative methods for estimating canopy height, such as stereo reconstruction from aerial imagery~\cite{aasen2015generating} or ground platforms~\cite{salas2017high}. Relying solely on canopy height for biomass prediction may lack effectiveness due to limited height variation in breeding programs. Beyond canopy height, other variables exhibit strong correlations with measured biomass, including point volume, LiDAR projected volume, and 3D indices~\cite{ jimenez2018high, walter2019estimating, sun2018field, caballer2021prediction, loudermilk2009ground}. Strategies centered around point density, which harnesses the inherent three-dimensional nature of point clouds, have been applied to diverse contexts and crops~\cite{sun2018field, greaves2015estimating, polo2009tractor, walter2019estimating}, such as cotton~\cite{ sun2018field}, arctic shrubs~\cite{ greaves2015estimating}, trees~\cite{polo2009tractor}, and wheat cultivated within specific environments~\cite{ walter2019estimating}. Noteworthy among these approaches is the voxel-based method (3DVI) introduced by Jimenez-Berni et al.~\cite{ jimenez2018high}, involving the division of point clouds into equal-sized voxels and subsequent calculation of the ratio between voxel-containing points and subdivisions within the horizontal plane. 
Significantly, the 3DVI technique, acknowledged as the 'gold standard', stands robust and precise for real-world applications, as evidenced by its implementation in various studies~\cite{ walter2019estimating, ten2019biomass}.

\subsubsection{Deep-learning-based biomass estimation}
Recent advancements in deep neural networks also showcased potential in biomass prediction. Oehmcke et al.~\cite{oehmcke2022deep} explored different neural network architectures for point cloud regression, yielding notable improvements in 
biomass estimates compared to conventional methods. Pan et al.~\cite{pan2022biomass} developed another deep-learning architecture, integrating completion, regularization, and projection modules alongside an attention-based fusion block for enhanced biomass predictions, exhibiting substantial enhancements over current state-of-the-art methods. Ma et al.~\cite{ma2023automated} proposed to use the dynamic graph CNN to predict the biomass with RGB-D camera data representation, showing a notable improvement over the 3DVI and Random forest methods. 

To benchmark the crop biomass prediction methods on MMCBE, we select the state-of-the-art as the baselines to evaluate the predicting accuracy. For the traditional density-based method, we chose the 3DVI~\cite{jimenez2018high, walter2019estimating} as one of the baselines. As for the deep learning-based approaches, BioNet~\cite{pan2022biomass} and DGCNN~\cite{ma2023automated} are also included in the baselines. The quantitative experimental results will be shown in the experimental section.

\begin{table}[tpb]
    \centering
    \caption{Comparison of existing datasets (including, but not limited to) on MMCBE. Modality includes LiDAR and Camera.}
    \label{table:datasets}
    \begin{adjustwidth}{-1.0cm}{}
    \scalebox{0.7}{
    \begin{tabular}{c|c|c c|c c|c}
        \hline
        \multirow{2}{*}{\textbf{Dataset}} &  \multirow{2}{*}{\textbf{Platform}} & \multicolumn{2}{c|}{\textbf{Modality}} & \multicolumn{2}{c|}{\textbf{Biomass estimation}} & \multirow{2}{*}{\textbf{Open access}} \\ 
        & & {\small\textbf{LiDAR}} & {\small \textbf{Camera}} & {\small \textbf{Traditional}}& {\small \textbf{Deep learning}} & \\ \hline
        \makecell{Jimenez et al.\\2018~\cite{jimenez2018high}} & UGV & \CheckmarkBold  & - & \CheckmarkBold & -  & -  \\ \hline
        \makecell{Wang et al.\\2019~\cite{wang2019mapping}} & UAV &  \CheckmarkBold & - &  \CheckmarkBold & - & - \\ \hline
        \makecell{Harkel et al.\\2019~\cite{ten2019biomass}}& UAV & \CheckmarkBold & - & \CheckmarkBold & -& - \\ \hline
        \makecell{Li et al.\\2020~\cite{li2020above}} & UGV & - & \CheckmarkBold & \CheckmarkBold& -& -  \\ \hline
        \makecell{Colaco et al.\\2021~\cite{colacco2021broadacre}} & UGV & \CheckmarkBold & - & \CheckmarkBold & -& - \\ \hline
        \makecell{Pan et al.\\2022~\cite{pan2022biomass}} & UGV &  \CheckmarkBold & - & -& \CheckmarkBold & \CheckmarkBold \\ \hline
        \makecell{Ma et al.\\2023~\cite{ma2023automated}} & \makecell{Static \\ support} & - & \CheckmarkBold & - & \CheckmarkBold & -  \\ \hline
        \makecell{Shu et al.\\2023~\cite{shu2023using}} & UAV  & - & \CheckmarkBold & \CheckmarkBold & - & -  \\ \hline
        \makecell{Our \\ MMCBE}& UAV & \CheckmarkBold & \CheckmarkBold & \CheckmarkBold & \CheckmarkBold &\CheckmarkBold  \\ \hline
    \end{tabular}}
    \end{adjustwidth}
\end{table}

\subsection{Existing crop datasets}
The availability and quality of datasets play a pivotal role in the advancement and application of deep learning algorithms, such as the importance of ImageNet~\cite{deng2009imagenet} and CoCo~\cite{lin2014microsoft} dataset to computer vision model. In the field of crop biomass prediction, there are very limited datasets, most of which are publicly unavailable. We provide a comparison between MMCBE with current datasets including, but not limited to these in Table~\ref{table:datasets}. LiDAR and cameras are still the main sensors for biomass phenotyping. LiDAR calculate the distance between the object and itself with time-to-flight and can generate accurate 3D geometrical structure about the object and scene, but its drawbacks are expensive and difficult to scale up to large fields, while cameras are widespread availabile and afford which are particularly appealing to biomass estimation. The UAV and unmanned ground vehicle (UGV) are the main vehicles to carry the sensor to capture the crop phenotyping data. With the improved capabilities of deep learning models, more deep learning-based approaches are applied in predicting crop biomass~\cite{pan2022biomass, ma2023automated}, in spite that conventional machine learning methods, such as linear regression and random forest, are still the mainstream methods. All these datasets only contain one modality, which is either LiDAR or Camera, and the majority of these datasets are not open-access, limiting their availability to the wider research community. Our dataset's use of dual sensor technology on a UAV platform for outdoor data collection and its open access policy potentially enhances its utility and applicability in advancing biomass prediction research. Apart from biomass crop prediction, the multi-view drone images can used for multiple computer vision topics, such as 3D reconstruction and novel-view synthesis.

\section{Multi-modality crop biomass dataset}
\label{sec:dataset}
\begin{figure}
    \centering
    \includegraphics[width=1.0\columnwidth]{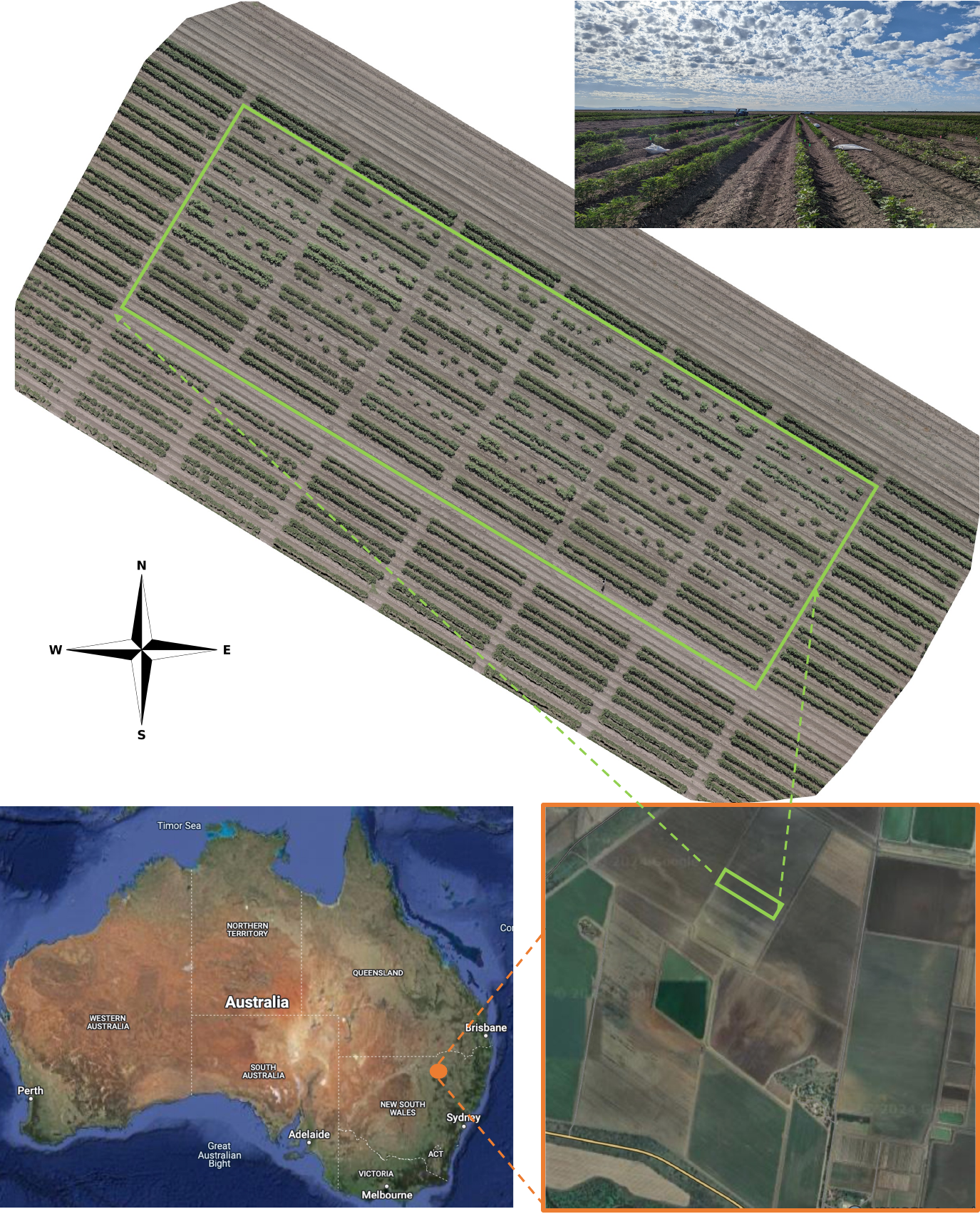}
    \caption{The experimental field for data collection.}
    \label{fig:location}
\end{figure}

\subsection{Field data collection}

\begin{figure*}[htb]
    \centering
    \includegraphics[width=1.8\columnwidth]{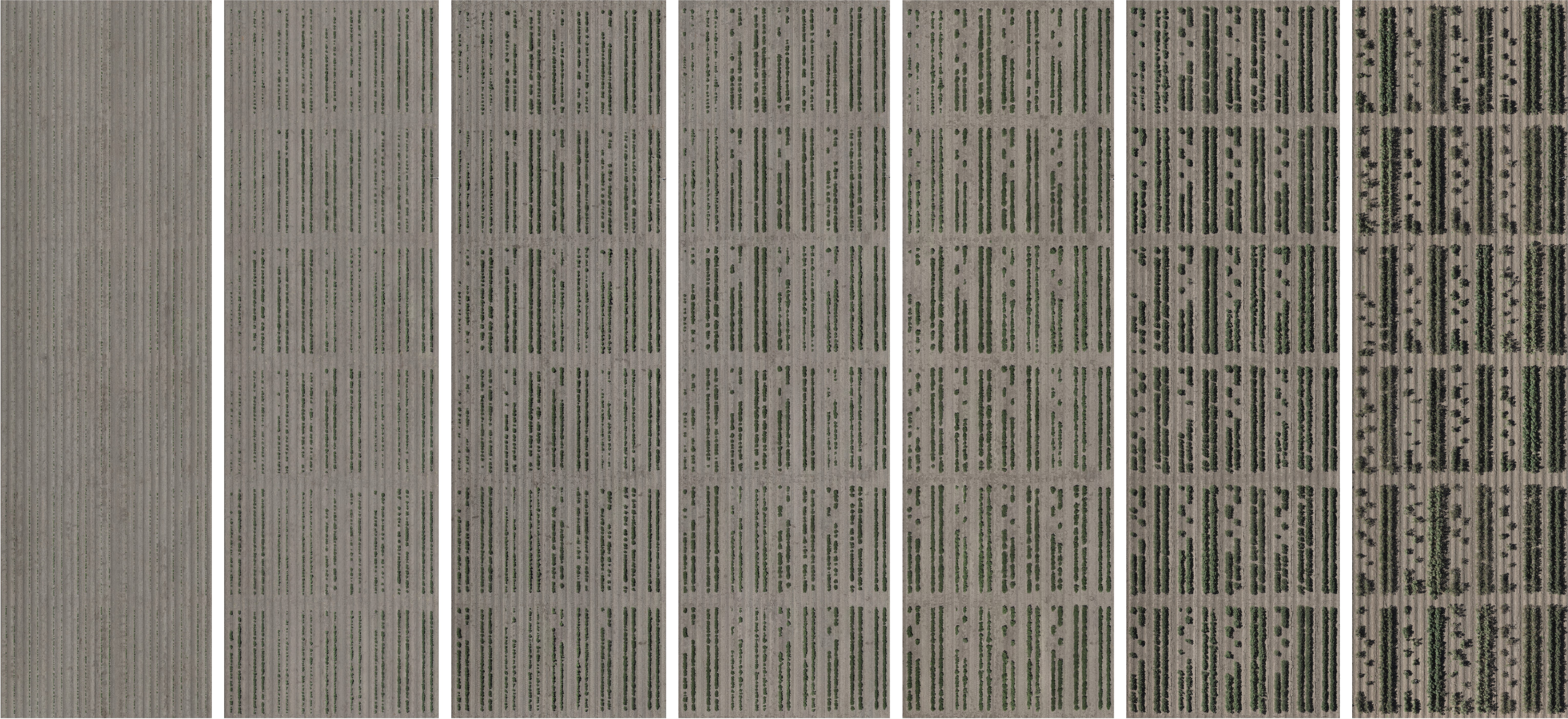}
    \caption{The visualization of how experimental fields change over time. The early to late growth stages of crops over the life cycle are presented from left to right. }
    \label{fig:temporal_change}
\end{figure*}

The dataset was collected from a dryland cotton field experiment grown at Australia, as shown in Figure~\ref{fig:location}. The experiment was sown on a row spacing of 1 m with a sowing density of ~10 plants ${m}^{-2}$. A 'single skip' row sowing configuration was used (i.e., two rows planted, one row skipped). Each plot consisted of two sets of two 13 m rows of cotton, with a 'skip' row in the middle of the plot and a shared skip row (with adjacent plots) on all borders of the plot. One set of two rows of cotton was used for non-destructive data collection, while the other set of two rows was used for destructive data collection.

\begin{figure}
    \centering
    \includegraphics[width=1.0\linewidth]{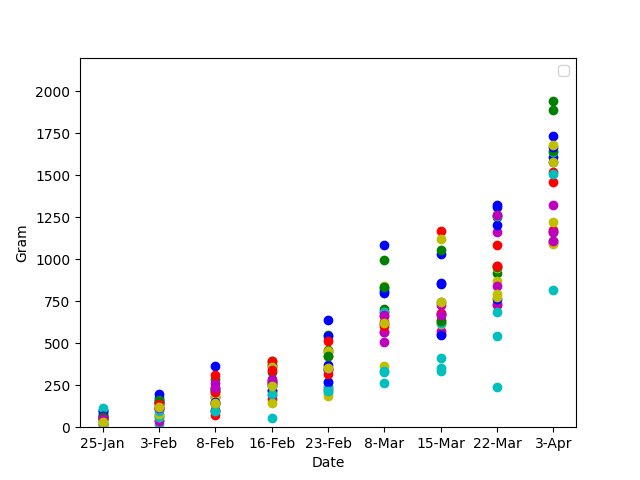}
    \caption{How manually measured above-ground biomass of each plot changes over time. For each time point, there are 24 sampling points, with four repetitions and six varieties. Each variety is marked with the same colour.}
    \label{fig:ground_truth}
\end{figure}

\begin{figure}[h!]
    \centering
    \includegraphics[width=1.0\linewidth]{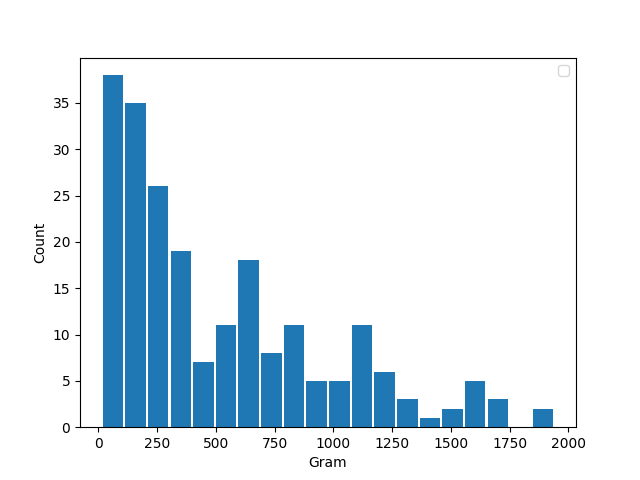}
    \caption{The histogram of the manually measured above-ground biomass.}
    \label{fig:histogram}
\end{figure}

\begin{figure*}[ht!]
    \centering
    \includegraphics[width=0.8\linewidth]{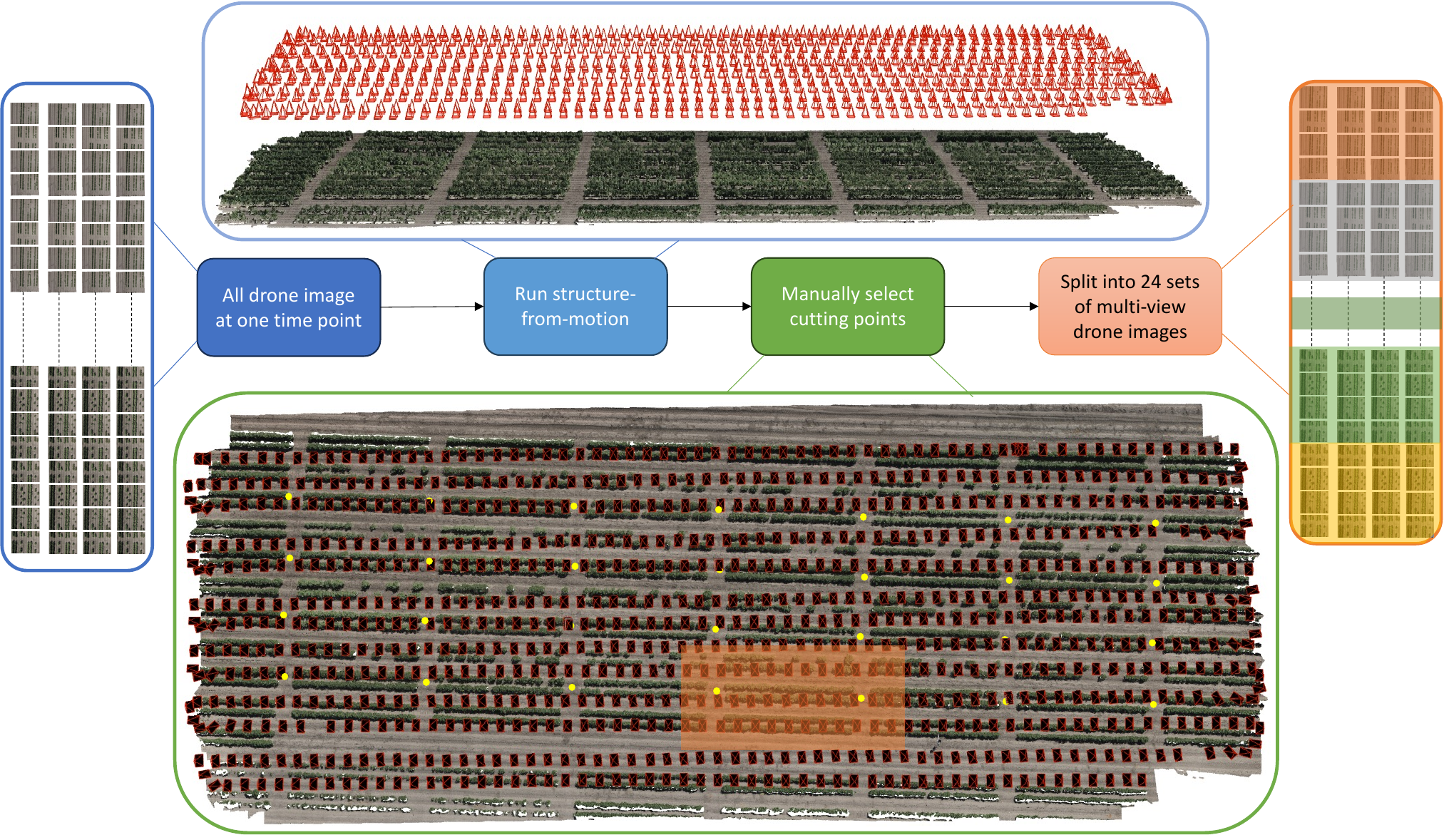}
    \caption{The processing pipeline of drone images: all drone images are input into SfM algorithms~\cite{schonberger2016structure} for 3D reconstruction and camera poses estimation, then we manually selected 28 points, marked with yellow dots, to locate each crop sample, and we can get drone multi-view images for each sample by calculating the distance between manually-selected points and camera poses. We finally crop 24 3D reconstructed point clouds and collect 24 multi-view image sets.}
    \label{fig:data_processing}
\end{figure*}

\begin{figure}[h!]
    \centering
    \includegraphics[width=0.85\linewidth]{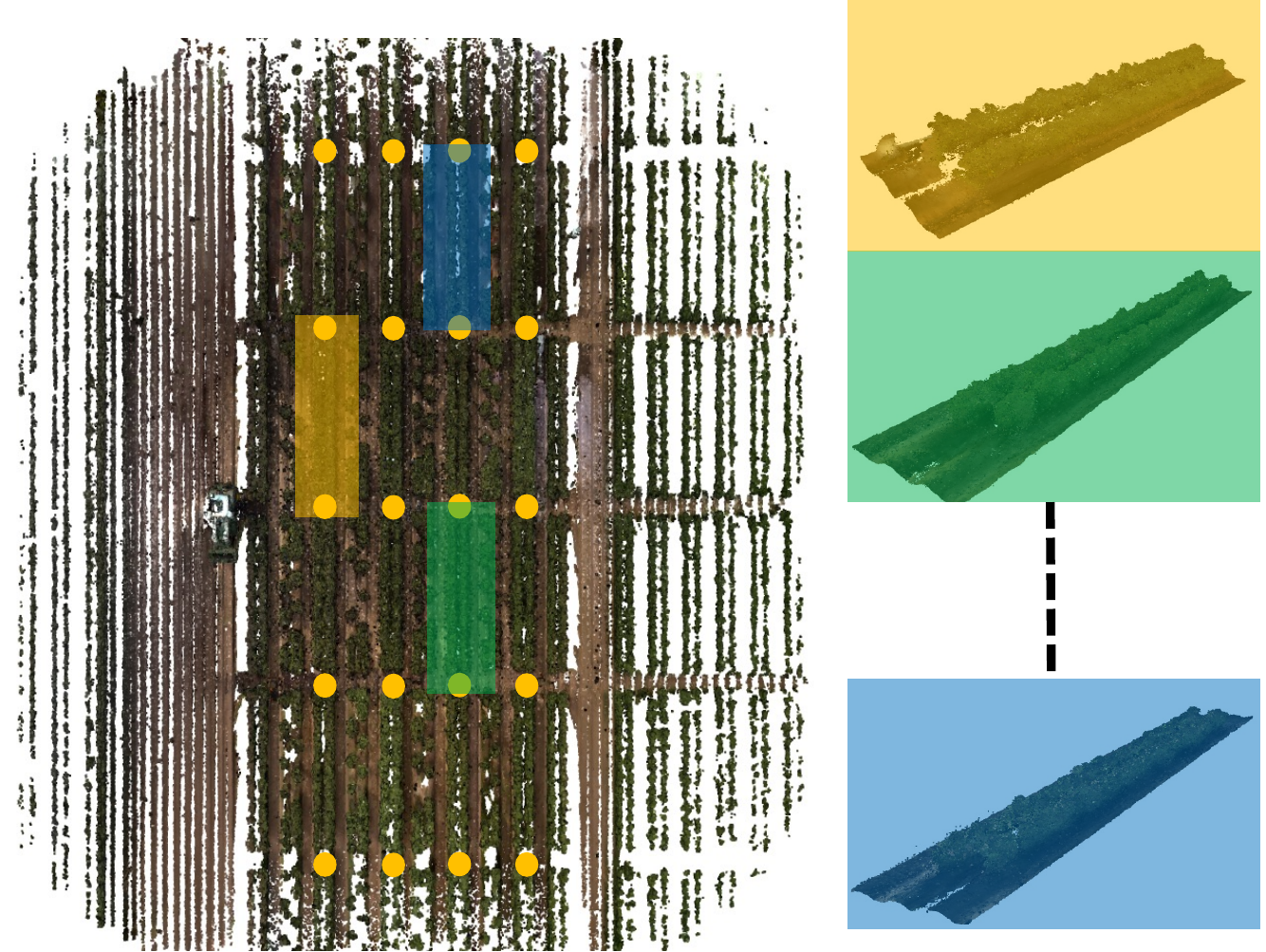}
    \caption{The visualization about how to generate LiDAR dataset.}
    \label{fig:lidar}
\end{figure}

The experimental area (0.216 ha) was scanned weekly by a P1 camera \footnote{https://enterprise.dji.com/zenmuse-p1} with the DJI M300 \footnote{https://enterprise.dji.com/matrice-300} flying 12 m above ground at a speed of 1.5 ${m}^{-s}$. The drone captured a new image for every 4 m travelled. Furthermore, we use the LiDAR from AutoMap\footnote{https://automap.io/terrusm/} to capture the point cloud of the plant on the same day. Biomass ground truth was collected on the day a flight was performed. We chose 9-time points, targeting key phenological plant stages during periods of high growth rate. The 9 time points are "25-Jan", "03-Feb", "08-Feb", "16-Feb", "23-Feb", "08-Mar", "15-Mar", "22-Mar", and "03-Apr". One ${m}^{2}$ of above ground plant material was cut at ground level from the 'destructive' data collection section of each plot. Plants were sampled from alternate rows up each plot to avoid any edge effects associated with reduced plant competition associated with previous biomass harvests. The number of plants and fresh weight were recorded. All sampled plants were then cut into ~50mm pieces and dried at 60 °C until no change in sample weight was observed ($\ge$ 48 h). Over the crop life cycle (6 months), we collected 216 (9 $\times$ 24) multi-view image sets with manually-measured biomass ground truth.

How measured above-ground biomass for each crop accumulates over time is shown in Figure~\ref{fig:ground_truth}, while the histogram of these biomass ground truths can be found in Figure~\ref{fig:histogram}. The OpenDroneMap\footnote{https://github.com/OpenDroneMap/ODM} is used to stitch all images in one flight together, shown in Figure~\ref{fig:temporal_change}, from which we can easily tell how crops grow over time in a whole field scale.





\begin{table*}[t]
    \centering
    \scalebox{1.0}{
    \begin{tabular}{c | c | c  c  c c c l} 
    \specialrule{.12em}{.12em}{.12em}
    Dataset & Modality & Metric& PointNet\cite{qi2017pointnet} & PoinNet++\cite{qi2017pointnet++} & DGCNN\cite{wang2019dynamic} & BioNet \cite{pan2022biomass} & 3DVI \cite{jimenez2018high}\\
    \specialrule{.12em}{.12em}{.12em}
     \multirow{6}{*}{\makecell{MMCBE \\(our dataset)}} & \multirow{3}{*}{\makecell{Image}} & MAE $\downarrow$ & 177.2  & 166.2  & 156.7  & \underline{153.8}  & 206.2  \\ 
      & & MARE $\downarrow$& 0.32  & 0.311  & \underline{0.283} & 0.292  & 0.428 \\ 
      & & RSME $\downarrow$     & 273.3  & 252.8  & 233.6  & \underline{228.4}  & 314.2 \\ \cline{2-8}
     & \multirow{3}{*}{\makecell{Point \\ cloud}} & MAE $\downarrow$ & 172.4 & 164.81 & 133.14 &  \underline{128.76}  &  130.34  \\ 
     & & MARE $\downarrow$& 0.253 & 0.229 & 0.236 &  \underline{0.201} & 0.225 \\ 
     & & RSME $\downarrow$ & 256.3 & 215.55 & 203.57 &  \underline{194.51} & 204.83 \\  \specialrule{.12em}{.12em}{.12em}

     \multirow{3}{*}{\makecell{Wheat \\dataset~\cite{pan2022biomass}}} & \multirow{3}{*}{\makecell{Point \\ cloud}} & MAE $\downarrow$ & 139.61 & 142.80 & 129.66 & \underline{71.23} & 115.15 \\ 
     & & MARE $\downarrow$& 0.274 & 0.275 & 0.254 & \underline{0.121} & 0.190 \\ 
     & & RSME $\downarrow$ & 189.80 & 188.85 & 161.61 & \underline{99.33} & 151.79 \\  \specialrule{.12em}{.12em}{.12em}
     
    \end{tabular}}
    \caption{Quantitative performance comparisons of SOTA for both point cloud and image modalities on our own dataset and another open-access dataset. The best results are marked are \underline{underlined}. For image modality, the input for SOTA methods is pseudo points obtained with SfM \cite{schonberger2016structure}.}
    \label{tab:exp_results}
\end{table*}

\subsection{Dataset processing}
When capturing the imagery data about crops, we fly the drone to collect images for the entire field (24 samples) in one go, therefore, we need to find the corresponding multi-view images for each sample. The data processing pipeline is illustrated with Figure~\ref{fig:data_processing}.

For each time point, we first run the structure-from-motion (SfM) algorithm~\cite{schonberger2016structure} to estimate camera poses and reconstruct the 3D structure. To find the images and reconstructed 3D point cloud for each sample, we manually selected its two ending points, as shown with yellow dots in Figure~\ref{fig:data_processing}. The sample's centre is calculated by averaging two ending points. 3D points of each sample will be cropped out, and we can find the distance from the sample's centre to all cameras. We can get the corresponding cameras' views for the sample based on the distance. Specifically, the directional norm $\vec{v}$ can be calculated for one sample with two ending points, meanwhile, the vectors from this sample's centre to all cameras' pose is easily obtained as $\vec{d}$. Then, we can calculate the distance from the camera to the sample's centre as $\vec{c} \cdot \Vec{d}$, and that from the camera to the $\vec{v}$ as $\vec{c} \times \Vec{d}$, we set the filtering distance to 1.5 and 7.5 meters for $\vec{c} \cdot \Vec{d}$ and $\vec{c} \times \Vec{d}$ respectively, the selected region is illustrated with a semi-transparent rectangle in Figure~\ref{fig:data_processing}. There exist GPS coordinates for each image, which can be an alternative to manually selecting points, but the GPS error may lead to wrong data association. After the correspondence between the crop sample and its associated camera views is found, the whole drone flight can be split into 24 multi-view image sets, each around 30 to 40 images.

Similarly, we manually select the two ending points for the LiDAR dataset, then calculate the distance of all points to the sample's centre and directional norm $\vec{v}$, based on which the sample's point cloud is cropped out. The LiDAR dataset can be found in Figure~\ref{fig:lidar}

\subsection{Challenges in our dataset}
\label{sec:challenges}
In contrast to datasets captured using UGV at a short sensing distance of 1 to 2 meters~\cite{pan2022biomass}, our dataset is acquired via a UAV at a considerably longer sensing distance from approximately 12 to 30 meters. This collection method introduces several challenges, outlines as follows:
\begin{itemize}
    \item The drone images are captured from forward-facing perspectives, which do not provide information on the lateral structure of crops. This limitation may introduce errors in biomass prediction methods that rely on associating 3D structures with biomass values.
    \item There is a low geometrical contrast between foreground and background. Given that the crops are typically less than 1 meter in height and substantially smaller than the sensing distance.
    \item The dataset mainly includes repetitive textures and exhibits a scarcity of structural features. It consists of only one type of crop, with similar appearances from a top-down perspective, adding complexity to 3D reconstruction efforts.
    \item Natural phenomena, such as winds, can cause inaccuracies in camera poses and lead to inconsistent representation of crop structure across different images. This inconsistency arises as we collect multi-view images by flying the UAV.
\end{itemize}

\noindent These challenges highlight the unique considerations required when using UAVs for data collection in agricultural settings and underscore the importance of developing robust methods to address these issues in biomass estimation and 3D reconstruction tasks.

\section{Exploration}
\label{sec:experiments}
Our dataset is mainly designed for the crop biomass prediction task, beyond this task, other computer vision tasks will be explored in our dataset as well. We first evaluate the performance of current state-of-the-art biomass prediction approaches on our dataset, and then we test the 3D reconstruction from multi-view images and novel-view synthesis.

 \begin{figure}[h]
    \centering
    \includegraphics[width=0.8\linewidth]{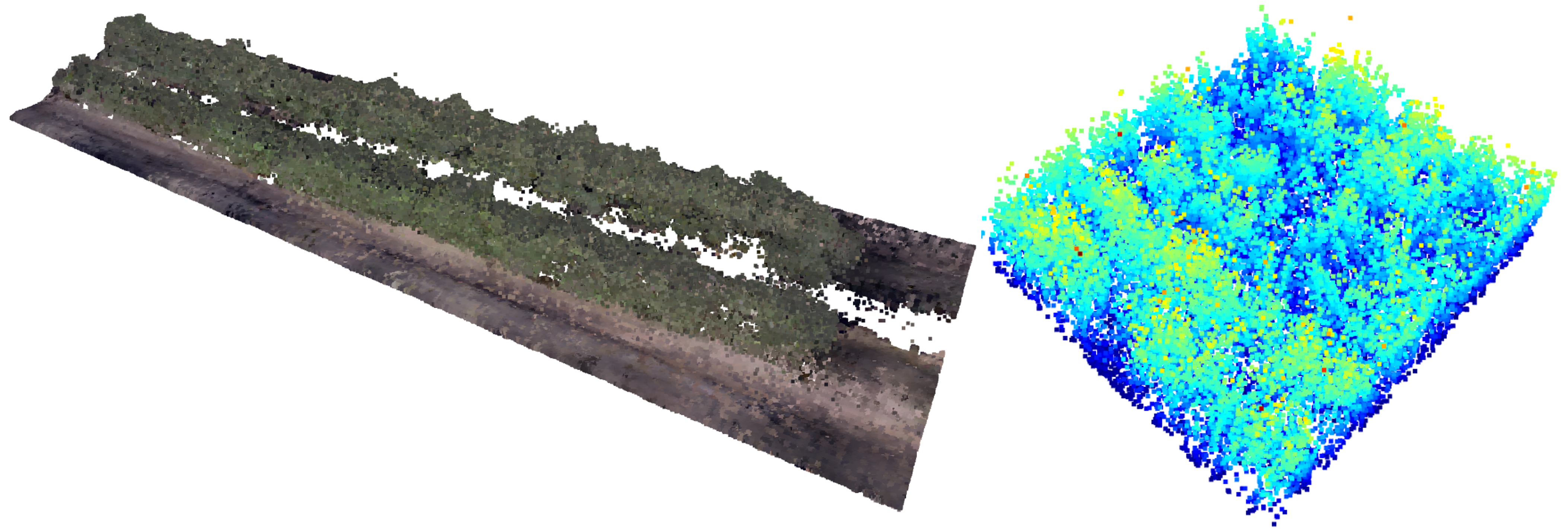}
    \caption{Visualization of a point cloud of publicly available wheat biomass dataset (right) and our self-collected dataset (left).}
    \label{fig:pc_vis}
\end{figure}

\subsection{Crop biomass estimation}
\subsubsection{Evaluation metrics}
We used similar metrics to evaluate biomass prediction errors as in \cite{pan2022biomass, ma2023automated}, namely mean absolute error \textit{(MAE)}, mean absolute relative error (\textit{MARE}), and root mean square error \textit{(RMSE}), as show in equation~\ref{equ:rmse}~\ref{equ:mae}~\ref{equ:mare}. \textit{MARE} offers the advantage of being more robust against the range of ground truth values observed across our 9 times points and different dataset.

\begin{equation}
    \text{RMSE} = \sqrt{\frac{1}{N} \sum_{i=1}^{N} (w_i - \hat{w}_i)^2}
    \label{equ:rmse}
\end{equation}

\begin{equation}
    \text{MAE} = \frac{1}{N} \sum_{i=1}^{N} |w_i - \hat{w}_i|
    \label{equ:mae}
\end{equation}

\begin{equation}
    \text{MARE} = \frac{1}{N} \sum_{i=1}^{N} \frac{|w_i - \hat{w}_i|}{w_i}
    \label{equ:mare}
\end{equation}

Where  \( N \) be the total point cloud data points with \( i \in \{1, \ldots, N\} \) as the index. Symbol \( \hat{w}_i \) is the predicted and \( w_i \) the ground-truth biomass value.

\begin{figure}
\begin{subfigure}{.48\linewidth}
  \centering
  \includegraphics[width=.98\linewidth]{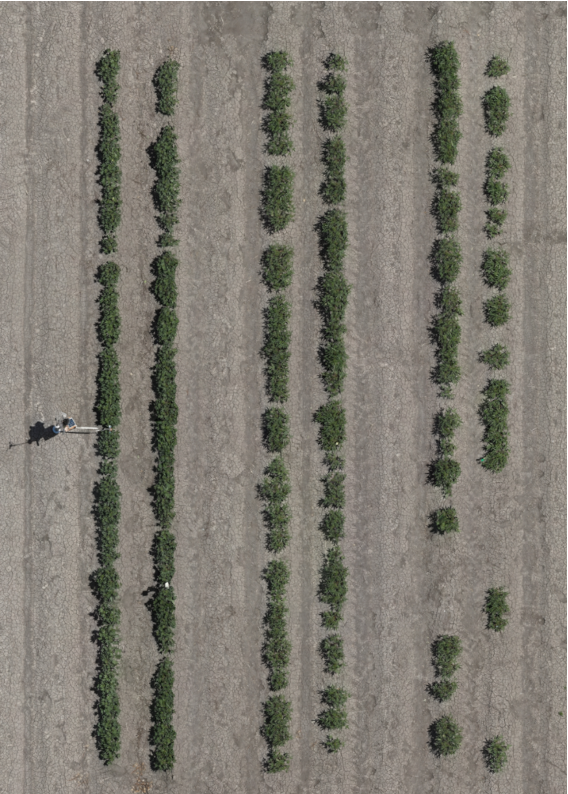}
  \caption{Orthophoto}
  \label{fig:sfig1}
\end{subfigure}%
\begin{subfigure}{.48\linewidth}
  \centering
  \includegraphics[width=.98\linewidth]{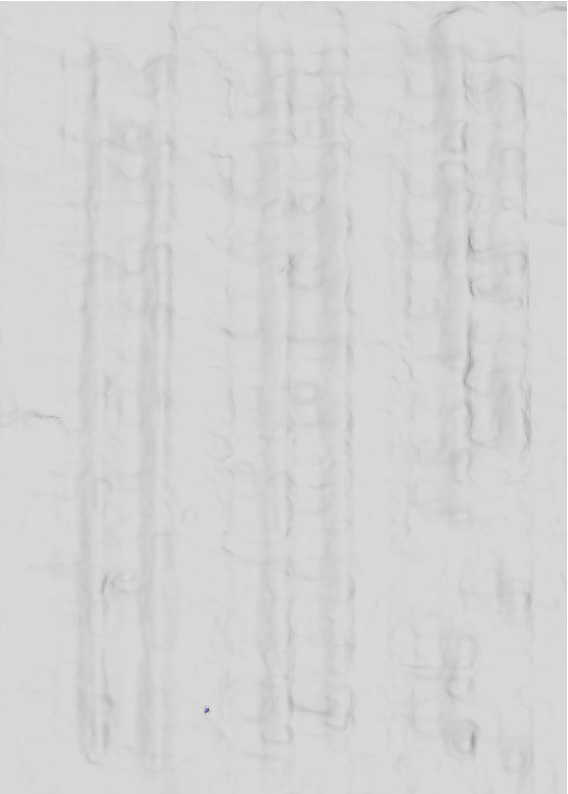}
  \caption{NeuS mesh}
  \label{fig:sfig2}
\end{subfigure}
\begin{subfigure}{.48\linewidth}
  \centering
  \includegraphics[width=.98\linewidth]{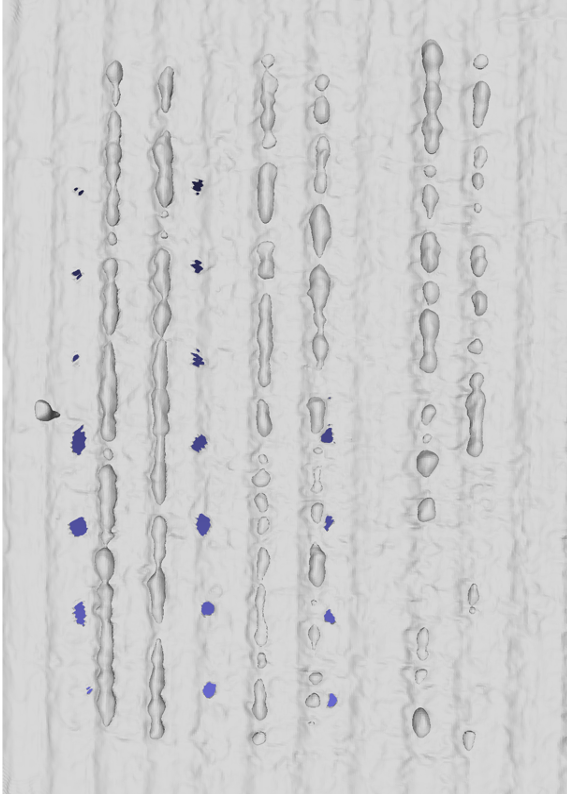}
  \caption{COLMAP mesh}
\label{fig:sfig3}
\end{subfigure}
\begin{subfigure}{.48\linewidth}
  \centering
  \includegraphics[width=.99\linewidth]{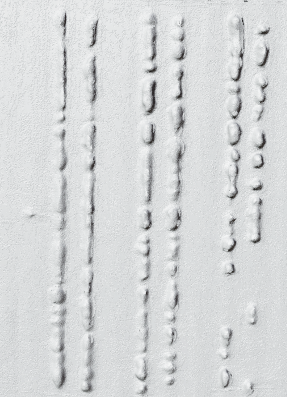}
  \caption{DUSt3R mesh}
\label{fig:sfig4}
\end{subfigure}

\begin{subfigure}{.48\linewidth}
  \centering
  \includegraphics[width=.98\linewidth]{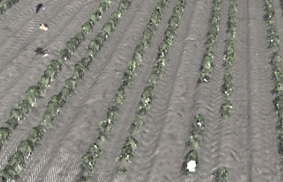}
  \caption{COLMAP mesh with color}
\label{fig:sfig3}
\end{subfigure}
\begin{subfigure}{.48\linewidth}
  \centering
  \includegraphics[width=1.\linewidth]{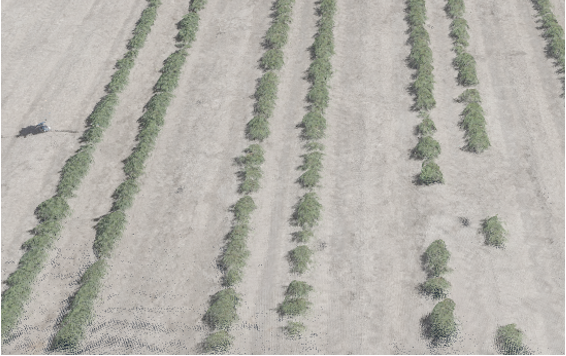}
  \caption{DUSt3R mesh with color}
\label{fig:sfig4}
\end{subfigure}

\caption{Visualization of three different multi-view 3D reconstruction approaches: NeuS~\cite{NeuS_3D_reconstru}, COLMAP~\cite{schonberger2016structure}, and DUSt3R~\cite{wang2023dust3r}.}
\label{fig:3D_reconstruction}
\end{figure}

\subsubsection{Comparison of current crop biomass prediction approaches}
 The existing SOTA methods, namely 3DVI~\cite{jimenez2018high, walter2019estimating}, BioNet~\cite{pan2022biomass}, PointNet~\cite{qi2017pointnet}, PointNet++~\cite{qi2017pointnet++}, and DGCNN~\cite{wang2019dynamic, ma2023automated}. Given that they all take point clouds as input, we used SfM~\cite{schonberger2016structure} to reconstruct point clouds from our multi-view drone images. Another public available dataset~\cite{pan2022biomass} is included for comparison as well, as shown in Tabel~\ref{tab:exp_results}. From the experimental table, we find that BioNet~\cite{pan2022biomass} consistently demonstrates superior performance across the majority of the metrics and modalities, particularly highlighted by its performance on the Wheat dataset, followed by DGCNN~\cite{wang2019dynamic, ma2023automated}. 
 
 When adopting the same approach, the point cloud modality exhibits higher prediction accuracy than the image modality. This distinction is likely attributed to the superior quality and more accurate geometrical structure of plants by LiDAR in contrast to the pseudo points generated by SfM~\cite{schonberger2016structure} from multi-view images. As for the same modality, i.e. point cloud, the same biomass prediction approach achieves better accuracy on wheat dataset~\cite{pan2022biomass} than our dataset. This could be caused by differences in point cloud quality and sample size, seen in Figure~\ref{fig:pc_vis}, the wheat dataset is collected with a close scanning distance from a UGV and has a more complete point cloud than ours. Moreover, size of each sample area for the wheat dataset is $1\times1~m^2$ while ours is $1\times15~m^2$. The large area tends to have a large biomass value, resulting in larger $MAE$ and $RSME$.

\subsection{Beyond crop biomass estimation}
As highlighted in Section~\ref{sec:challenges}, our dataset encompasses scenarios that present significant challenges. To investigate our dataset's applicability and potential contributions to other computer vision tasks beyond the primary scope of biomass prediction, we explore 3D reconstruction and novel-view rendering tasks in the subsequent sections.

\subsubsection{3D reconstruction from crops}
This sub-section explores how well the current 3D reconstruction approaches perform on our multi-view drone images. Three different methods, i.e. COLMAP~\cite{schonberger2016structure}, NeuS~\cite{NeuS_3D_reconstru}, and DUSt3R~\cite{wang2023dust3r}, are chosen for evaluation. The COLMAP~\cite{schonberger2016structure} is a typical conventional 3D reconstruction using structure-from-motion, it first extracts and matches key points across multiple images (seeing Figure~\ref{fig:sceenshot}), then estimates the camera poses with bundle adjustment, and finally performs dense reconstruction. NeuS~\cite{NeuS_3D_reconstru} represents the NeRF-based 3D reconstruction method by introducing a signed distance function (SDF) into NeRF~\cite{mildenhall2021nerf} optimization framework, and 3D reconstruction can be completed with RGB image supervision only, and the scene's mesh can be extracted from SDF. Different from the other two methods, DUSt3R~\cite{wang2023dust3r} is a stereo 3D reconstruction from un-calibrated and un-posed cameras, its network can regress a dense point cloud solely from a pair of images, without prior information about the scene and camera. The qualitative results are shown in Figure~\ref{fig:3D_reconstruction}, from which we can find that NeuS~\cite{NeuS_3D_reconstru} fails to do 3D reconstruction in our dataset, one possible reason is that our scene has the repetitive texture and the plants are quite small and also close to this soil background. DUSt3R can reconstruct 3D plant roughly and smoothly, but its reconstructed ground is very flat which is inaccurate as we know that the experimental fields have waving curves for irrigation and sowing. COLMAP can reconstruct both 3D plants and the ground's geometry. In summary, the conventional structure-from-motion approaches are better for 3D reconstruction tasks in our dataset due to our challenging scenarios and the imitated presence of a dataset similar to ours. In our opinion, the prior knowledge about drone scenarios can be introduced into NeuS~\cite{NeuS_3D_reconstru} to improve 3D reconstruction from multi-view drone images, while DUSt3R~\cite{wang2023dust3r} can be further boosted for our scenarios by fine-tuning model with drone image dataset.

\subsubsection{Novel view rendering}
\begin{figure}
    \centering
    \includegraphics[width=9cm,height=5cm]{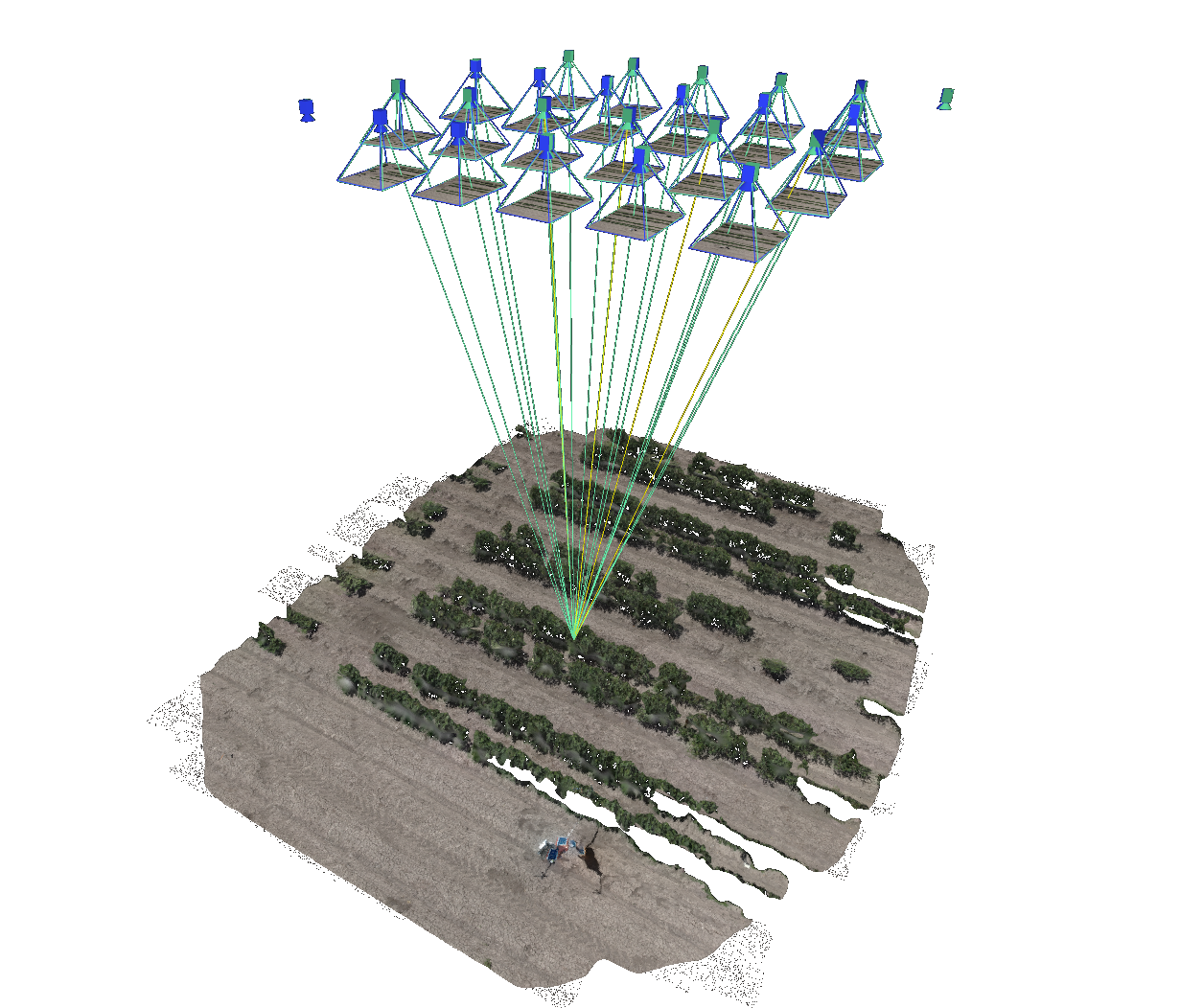}
    \caption{Illustration of a light ray for different images in NeRF, meanwhile it can be used to demonstrate 3D reconstruction by matching 3D key points.}
    \label{fig:sceenshot}
\end{figure}

\begin{figure}
    \includegraphics[width=.85\linewidth]{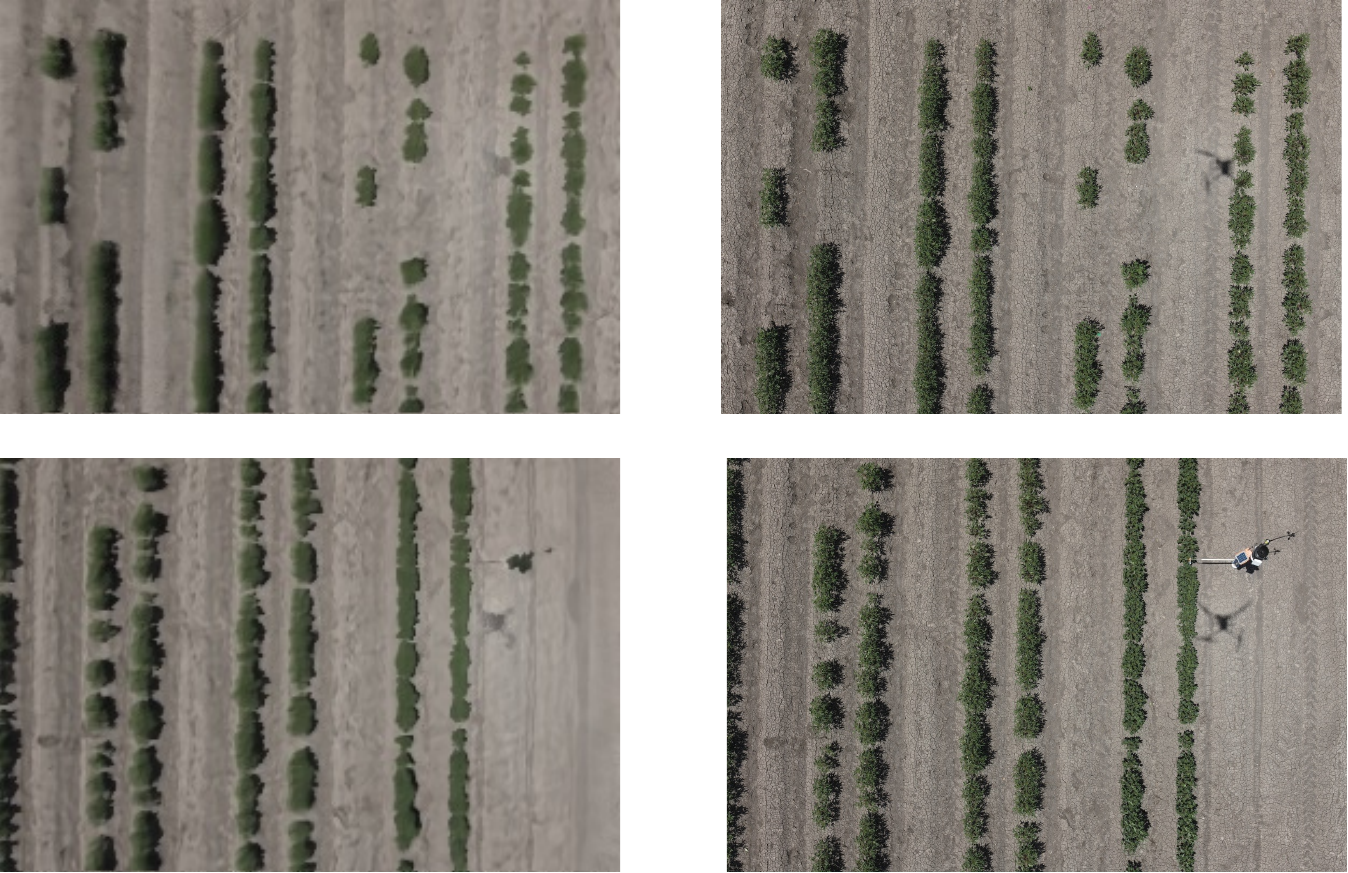}
\caption{Visualization of two rendered images. The left two images are rendered one, and the right two images are the corresponding real one.}
\label{fig:rendering}
\end{figure}

Novel view rendering can generate images of crop scenes from specific viewpoints. Among the various methods available, NeRF~\cite{mildenhall2021nerf} has gained popularity for this purpose by learning a neural implicit field for 3D scene representation. In NeRF, each pixel in the image is considered as a light ray with its colour value being a weighted sum of all radiance values (extracted from the neural radiance field) along the ray, as shown in Figure~\ref{fig:sceenshot}. This approach holds the potential for selecting the most crucial key views from drone-captured images, which in turn can help optimize drone flight trajectories and sampling rates. In this section, we assess the applicability of this task to our dataset. The quantitative outcomes are presented in Table~\ref{tab:PSNR}, while qualitative results are shown in Figure~\ref{fig:rendering}. These results indicate that, within our dataset, NeRF achieves decent novel-view rendering quality, albeit not high-fidelity rendering. This limitation may stem from specific challenges posed by our scenarios such as forward-facing perspectives, a scarcity of structural features, and inaccurate camera poses (mentioned in Section~\ref{sec:challenges}).

\begin{table}[h!]
    \centering
    \scalebox{1.1}{
    \begin{tabular}{c|c|c|c|c|c} 
    \specialrule{.1em}{.1em}{.1em}
    Scene ID & \#1 & \#3 & \#5 & \#7 & \#9 \\ \specialrule{.1em}{.1em}{.1em}
    \centering PSNR$\uparrow$ & 25.41 & 23.31 & 21.92 & 21.6 & 22.54  \\\hline
    \centering SSIM$\uparrow$ & 0.90 & 0.89 & 0.82& 0.82 & 0.85 \\\hline
    \end{tabular}}
    \caption{Quantitative performance of NeRF~\cite{mildenhall2021nerf} rendering novel-views on 1st, 3rd, 5th, 7th, and 9th time points. Rendering results of all image sets in a single time point were averaged.}
    \label{tab:PSNR}
\end{table}

\section{Conclusion}
\label{sec:conclusion}

We introduced a new multi-modality dataset in this paper, aiming to further innovation and research in accurate, low-cost, and scalable biomass estimation methods. The dataset includes point clouds and images with manually-labelled ground truth. We evaluated current state-of-the-art methods on our dataset for crop biomass prediction, demonstrating its capability to provide more training data and serve as a transparent benchmark for performance evaluation. Additionally, we explored its application in other computer vision tasks. It is noteworthy that multi-view-based 3D reconstruction and novel-view rendering tasks faced challenges due to the unique properties of our dataset, underscoring its potential value to the broader computer vision field.


{
    \small
    \bibliographystyle{IEEEtran}
    \bibliography{main}
}


\end{document}